\documentclass{article}

\usepackage{microtype}
\usepackage{graphicx}
\usepackage{subfigure}
\usepackage{booktabs} 

\usepackage{hyperref}

\usepackage[accepted]{icml2021}

\usepackage{amsmath}
\usepackage{lipsum}

\usepackage{setspace}
\usepackage[moderate]{savetrees}
\makeatletter
\renewcommand\section{\@startsection{section}{1}{\z@}{-0.06in}{0.001in} {\large\bf\raggedright}}
\def\subsection{\@startsection{subsection}{2}{\z@}{-0.05in}{0.001in}{\normalsize\bf\raggedright}}
\renewcommand\subsubsection{\@startsection{subsubsection}{3}{\z@}{-0.04in}{0.0001in}  {\normalsize\bf\raggedright}}
\makeatother

\usepackage[table]{xcolor}
\usepackage{multirow}
\usepackage{makecell}
\usepackage{arydshln}
\usepackage{colortbl}
\definecolor{Gray}{gray}{0.86}
\definecolor{LightGray}{gray}{0.94}
\usepackage{xcolor}
\definecolor{darkred}{HTML}{bb0000}
\definecolor{darkpink}{HTML}{ca00ff}
\renewcommand\ttdefault{cmtt}

\usepackage{amsmath,amsfonts,bm}

\renewcommand{\slash}{\!/\! }
\newcommand{\heading}[1]{\textbf{#1}~~}
\newcommand{\eg}{\textit{e.g.}}
\newcommand{\ie}{\textit{i.e.}}
\newcommand{\trainbg}{\textsc{Train}^{\text{bad}\rightarrow \text{good}}}
\newcommand{\traingb}{\textsc{Train}^{\text{good}\rightarrow \text{bad}}}

\def\eqref#1{equation~\ref{#1}}

\def\1{\bm{1}}

\DeclareMathAlphabet{\mathsfit}{\encodingdefault}{\sfdefault}{m}{sl}
\SetMathAlphabet{\mathsfit}{bold}{\encodingdefault}{\sfdefault}{bx}{n}

\def\gD{{\mathcal{D}}}

\def\gP{{\mathcal{P}}}

\usepackage{mathtools}

\icmltitlerunning{Break-It-Fix-It: Unsupervised Learning for Program Repair}

\begin{document}
\setlength{\abovedisplayskip}{6pt}
\setlength{\belowdisplayskip}{6pt}

\twocolumn[
\icmltitle{Break-It-Fix-It: Unsupervised Learning for Program Repair}



\icmlsetsymbol{equal}{*}

\begin{icmlauthorlist}
\icmlauthor{Michihiro Yasunaga}{su}
\icmlauthor{~~Percy Liang}{su}
\icmlauthor{~~~~~~~}{}
\end{icmlauthorlist}

\icmlaffiliation{su}{Stanford University, Stanford, CA}

\icmlcorrespondingauthor{Michihiro Yasunaga}{myasu@cs.stanford.edu}

\icmlkeywords{Machine Learning, ICML}

\vskip 0.3in
]



\printAffiliationsAndNotice{}  

\begin{abstract}
We consider repair tasks: given a \textit{critic} (\eg, compiler) that assesses the quality of an input, the goal is to train a fixer that converts a bad example (\eg, code with syntax errors) into a good one (\eg, code with no syntax errors). 
Existing works create training data consisting of (bad, good) pairs by corrupting good examples using heuristics (\eg, dropping tokens).  However, fixers trained on this synthetically-generated data do not extrapolate well to the real distribution of bad inputs.
To bridge this gap, we propose a new training approach, \textit{Break-It-Fix-It} (\textit{BIFI}), which has two key ideas:
(i) we use the critic to check a fixer's output on real bad inputs and add good (fixed) outputs to the training data,
and (ii) we train a \textit{breaker} to generate realistic bad code from good code. 
Based on these ideas, we iteratively update the breaker and the fixer while using them in conjunction to generate more paired data.
We evaluate BIFI on two code repair datasets: GitHub-Python, a new dataset we introduce where the goal is to repair Python code with AST parse errors; and DeepFix, where the goal is to repair C code with compiler errors. BIFI outperforms state-of-the-art methods, obtaining 90.5\% repair accuracy on GitHub-Python (+28.5\%) and 71.7\% on DeepFix (+5.6\%).
Notably, BIFI does not require any labeled data; we hope it will be a strong starting point for unsupervised learning of various repair tasks.
\vspace{-4mm}
\end{abstract}
\section{Introduction}
\begin{figure}[!t]
    \vspace{-1mm}
    \hspace{-2mm}
    \includegraphics[width=0.49\textwidth]{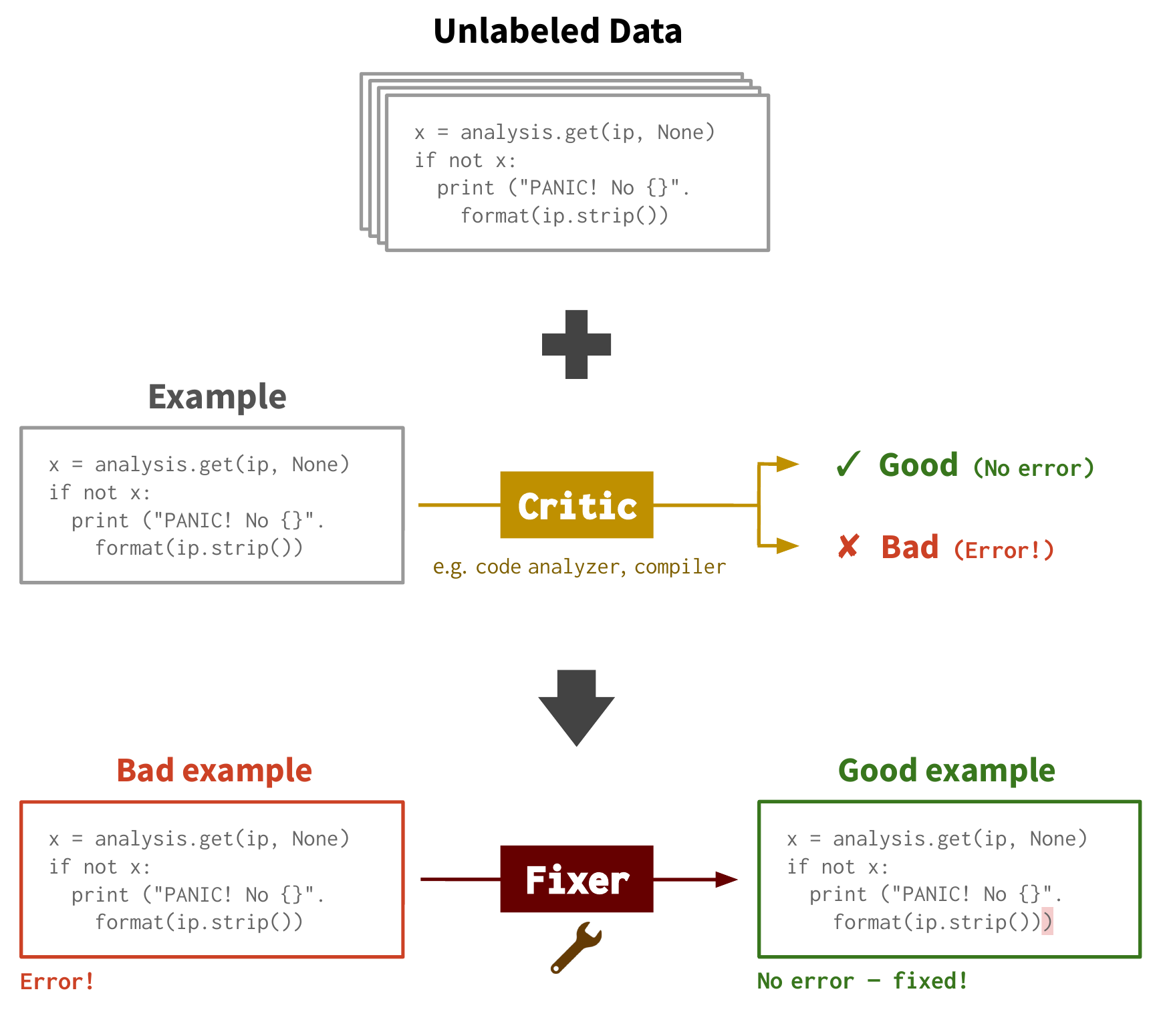}\vspace{-7mm}
    \caption{
    \textbf{Problem setup}. We are given an \textbf{unlabeled data} (code snippets) and a \textbf{critic} (\eg, code analyzer, compiler) that assesses the quality of an input (\eg, bad if the code has errors; good if no error).
    Our task is to learn a \textbf{fixer} that can actually repair a bad example into a good one (\eg, fixing errors in the bad code).
    } \vspace{-2mm}
  \label{fig:task}
\end{figure}

\begin{figure}[!h]
    \vspace{-1mm}
    \hspace{-1mm}
    \includegraphics[width=0.49\textwidth]{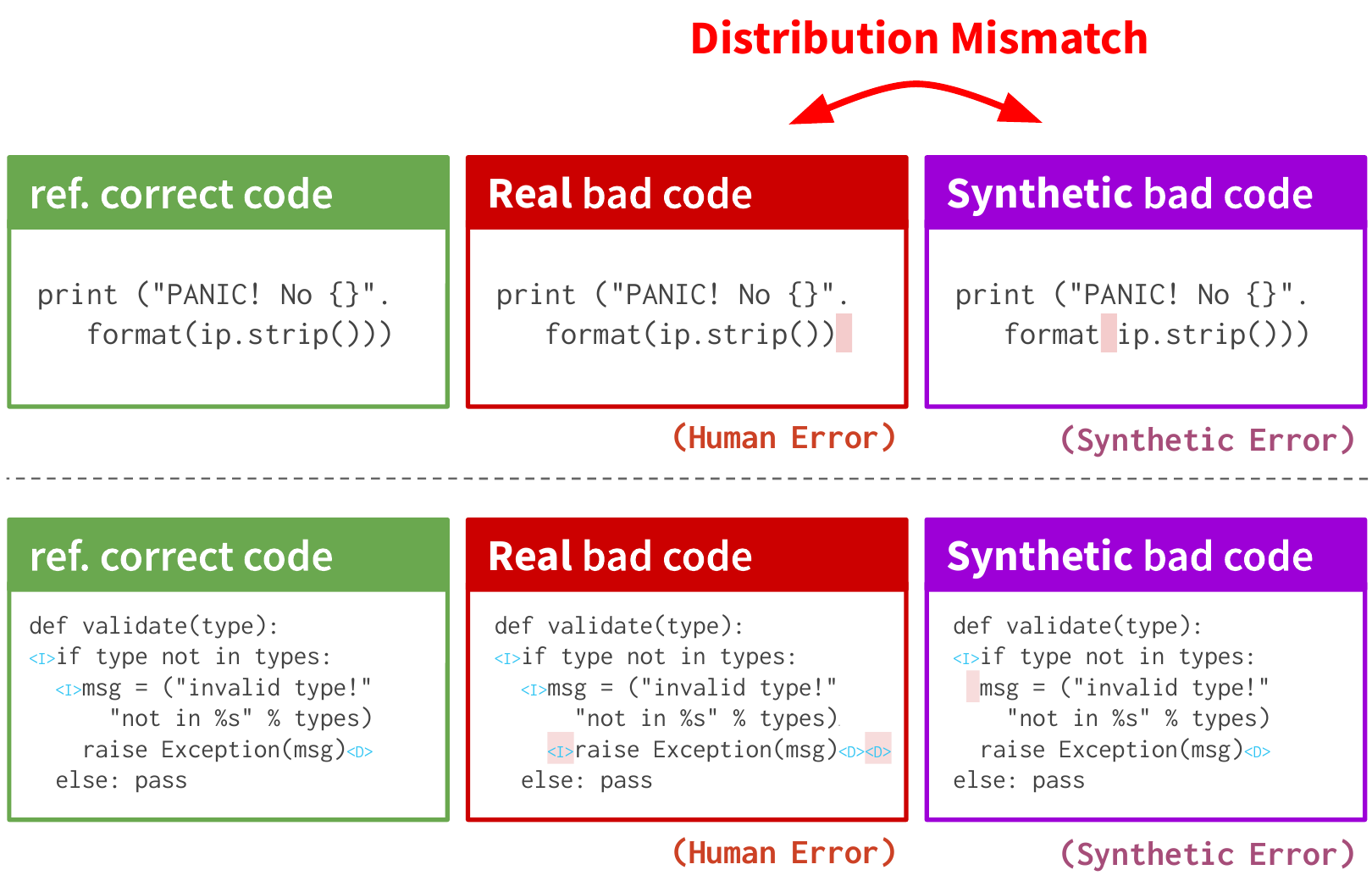}\vspace{-6mm}
    \caption{
    \textbf{Challenge of learning a fixer from unlabeled data}. Existing works randomly perturb good examples into bad examples and learn to recover. However, such {synthetically} generated bad examples do not match the distribution of \textit{real} bad examples. 
    For instance, synthetic perturbations may drop parentheses arbitrarily from code (top right), but real human-written bad code misses parentheses more often in a nested context (top center).
    In a more extreme case at bottom, to generate the human error (center) from the corrected code (left),
    a \textit{pair} of indent and dedent tokens need to be inserted accordingly, which random perturbations generate with very small probability.
    Note that {indentation is meaningful in Python}: in the tokenized Python code, each \texttt{$\langle$I$\rangle$} token means indenting the line by one unit, each \texttt{$\langle$D$\rangle$} means dedenting the next line by one unit.}\vspace{-4mm}
  \label{fig:challenge}
\end{figure}

\begin{figure*}[!t]
    \centering
    \includegraphics[width=0.97\textwidth]{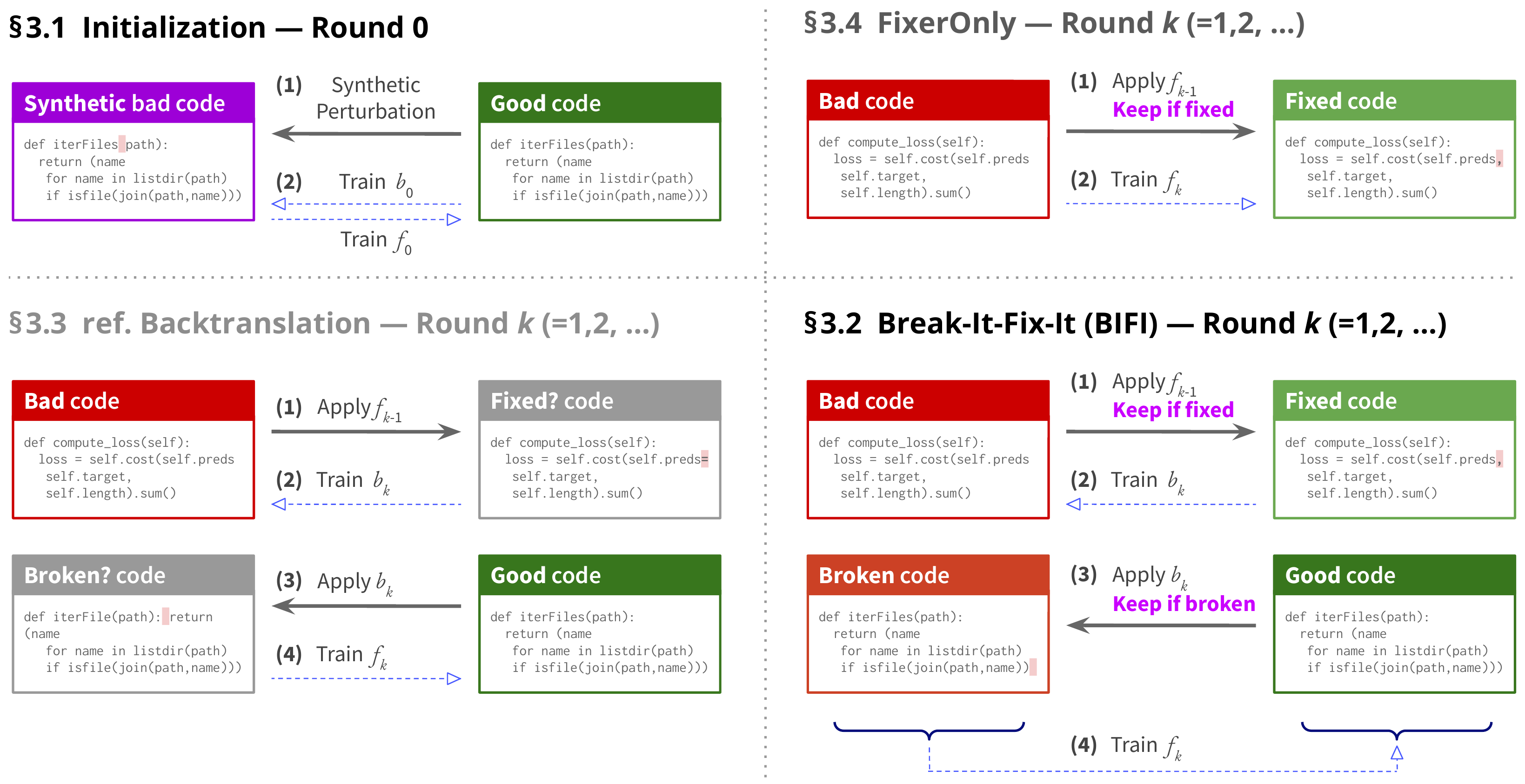}\vspace{-3mm}
    \caption{
    \textbf{Overview of our approach}. We train the initial fixer on synthetically prepared data as in prior works (\textbf{top left}). 
    In our approach, \textit{BIFI} (\textbf{bottom right}), 
    we apply the initial fixer to the real bad examples and add fixed outputs (verified by the critic) to our training data (Step 1), train a breaker on the resulting paired data (Step 2), use the breaker to generate (intuitively more realistic) code errors (Step 3), and train the fixer again on the newly-generated paired data (Step 4). We iterate this cycle, improving the fixer and the breaker simultaneously.
    \textbf{Top right}: a version of BIFI without the breaker (\textit{FixerOnly}). \textbf{Bottom left}: comparison of BIFI to backtranslation. The main difference is that BIFI uses the critic to verify that the fixer produces good code and the breaker produces bad code (\textcolor{darkpink}{\textbf{annotated with magenta font}}).
    }\vspace{-2mm}
  \label{fig:approach}
\end{figure*}

In many domains, one has access to a \textit{critic} that assesses the quality of an input, but what is desired is a more constructive \textit{fixer} that actually improves bad inputs. For instance, in programming, while a code analyzer and compiler can tell if a given code has errors, programmers need to repair the errors in the bad code. Development of an automatic code fixer is thus a key to enhancing programming productivity \cite{seo2014program} and is an active area of research \cite{mesbah2019deepdelta,ding2020patching,dinella2020hoppity}.
Other instances of this general setting include molecular design \cite{jin2018learning} which aims to improve the chemical properties (\eg, drug-likeness) of molecules given a property evaluator, and essay editing \cite{taghipour2016neural} which aims to improve a writing given a grade. How to automatically learn a fixer given a critic (we term \textit{critic2fixer}) remains an important research problem in machine learning.

In this work, we focus on the domain of code repair. Learning a fixer is challenging because manual labeling of paired data, \eg, $\langle$broken code, fixed code$\rangle$, is costly.
To this end, we consider learning from unlabeled data. Specifically, as illustrated in Figure \ref{fig:task},
we are given (a) a \textbf{critic} (code analyzer or compiler) that assesses the quality of an input---\textit{bad} if the code has errors; \textit{good} if it has no errors---, and (b) \textbf{unlabeled data}---unpaired set of good code and bad code, \eg, from GitHub. Our goal is to learn a \textbf{fixer} that repairs bad code into good code.
Previous works in code repair apply random or heuristic perturbations to good code (\eg, dropping tokens) and prepare synthetic paired data $\langle$perturbed code, good code$\rangle$ to train a fixer \cite{pu2016sk_p,gupta2017deepfix,ahmed2018compilation,hajipour2019samplefix,yasunaga2020repair}. However, such synthetically generated bad examples do not match the distribution of \textit{real} bad examples. For instance, as shown in Figure \ref{fig:challenge}, synthetic perturbations may drop parentheses arbitrarily from code, generating errors that rarely happen in real programming (Figure \ref{fig:challenge} top right; synthetic errors); in contrast, real human-written code misses parentheses more often in a nested context (Figure \ref{fig:challenge} top center; human errors). 
As we will show in \S \ref{sec:experiments}, this distribution mismatch between synthetic data and real data results in low performance.

To bridge this gap, we propose \textit{Break-It-Fix-It} (\textit{BIFI}), a new method to learn a fixer from unlabeled data and a critic (Figure \ref{fig:approach}). 
BIFI is based on two core insights: 
(i) we can use the critic to check a fixer's output on real bad examples and add good outputs to the training data,
and (ii) we train a \textit{breaker} to generate realistic bad examples from good examples.
Specifically, given an initial fixer trained on synthetic paired data $\langle$synthetic bad, good$\rangle$, BIFI improves the fixer and breaker simultaneously through rounds of data generation and training: (1) apply the fixer to real bad examples and keep fixed outputs to obtain real paired data, (2) use the resulting data to train the breaker, (3) use the learned breaker to generate code errors and obtain more paired data, and (4) train the fixer on the newly-generated paired data in (1) and (3).
Intuitively, this cycle trains the fixer on increasingly more real or realistically generated bad code, adapting the fixer from the initial synthetic distributions towards real distributions of code errors.

The BIFI algorithm is related to backtranslation in unsupervised machine translation \cite{lample2017unsupervised}, 
which uses a target-to-source model to generate noisy sources and trains a source-to-target model to reconstruct the targets (\eg, the bad-side and good-side in our repair task can be viewed as the source and target).
BIFI differs from backtranslation in two ways: it uses the critic to verify if the generated examples are actually fixed or broken (step 1 and 3), and it trains the fixer on \textit{real} bad examples in addition to examples generated by the breaker (step 4), which improves the correctness and distributional match of generated paired data.

We evaluate our proposed approach on two code repair datasets:
\begin{itemize}
\renewcommand\ttdefault{cmtt}
    \setlength{\leftskip}{2mm}
    \setlength{\itemsep}{1mm}
    \item[$\bullet$] \textbf{GitHub-Python}: We collected a new dataset of 3M Python code snippets from \href{https://github.com/}{\texttt{github.com}}.
    The task is to repair errors caught by the Python AST parser. We set the initial fixer to be an encoder-decoder Transformer \cite{vaswani2017attention} trained on random perturbations.
    \item[$\bullet$] \textbf{DeepFix} \citep{gupta2017deepfix}: The task is to repair compiler errors in C code submitted by students in an introductory programming course. We set the initial fixer to be the existing best system, DrRepair \cite{yasunaga2020repair}, which was trained on manually-designed heuristic perturbations.
\end{itemize}
Our approach (BIFI) outperforms the initial fixers, obtaining 90.5\% repair accuracy on GitHub-Python (+28.5\% absolute) and 71.7\% repair accuracy on DeepFix (+5.6\% absolute), attaining a new state-of-the-art. 
BIFI also improves on backtranslation by 10\%. Further, we qualitatively show how the fixer and breaker adapt towards more realistic distributions of code through the BIFI algorithm (\S \ref{sec:analysis}).

\section{Problem statement}
\label{sec:problem}

Figure \ref{fig:task} illustrates our problem setup, critic2fixer. 
The system is given unlabeled data $\gD$ (code snippets) and a {critic} $c$ (code analyzer or compiler) that returns whether an input is good or bad: \eg, for a code snippet $x\in \gD$,
\begin{align}\vspace{-1mm}
    c(x) = \begin{cases}
    0 ~~~\text{if $x$ has errors},\\[-1mm]
    1 ~~~\text{if $x$ has no error}.
    \end{cases}
    \vspace{-1mm}
\end{align}
Using the critic $c$, examples in $\gD$ can be classified into {bad} ones $\gD_\text{bad} = \{x \mid x \in \gD,~ c(x) = 0 \}$ and {good} ones 
$\gD_\text{good} = \{y \mid y \in \gD,~  c(y) = 1 \}$.
Our task is to learn a {fixer} $f$ that maps a bad example $x\in\gD_\text{bad}$ into a good example $f(x)$ such that it is close\footnote{We constrain the edit distance as described in \S \ref{sec:exp-setup}. We acknowledge that while we want $f(x)$ to be semantics-preserving, it is non-trivial to ensure this automatically, so we rely on the edit distance.} to $x$ and $c(f(x)) = 1$.
The evaluation metric is the fixer $f$'s repair accuracy on a held-out set of bad examples, $\gD_\text{bad}^\text{(test)}$,
\begin{align}
    \text{RepairAcc} = \frac{|\{x \mid x\in \gD_\text{bad}^\text{(test)}, ~c(f(x)) = 1\}|}{|\gD_\text{bad}^\text{(test)}|}.
\end{align}

\section{Approach}
\label{sec:approach}

The major challenge of learning a fixer is that we need to learn from unpaired data, \ie, $\gD_\text{bad}$ and $\gD_\text{good}$ do not form $\langle$broken, fixed$\rangle$ pairs.  Prior works in code repair apply random or heuristic perturbations to good examples (\eg, dropping tokens) and prepare a synthetic paired data $\langle$perturbed code, good code$\rangle$ to train a fixer \cite{gupta2017deepfix,ahmed2018compilation,yasunaga2020repair}. However, such synthetically generated bad examples do not match the distribution of real bad examples. 
For instance, as Figure \ref{fig:challenge} (top) shows, synthetic perturbations may drop parentheses arbitrarily from code, generating errors that are rare in real programs;
in contrast, real human-written code misses parentheses often in a nested context
(\eg, 10x more than non-nested in our collected dataset GitHub-Python).
In a more extreme case (Figure \ref{fig:challenge} bottom), to make the real human error (center) from the corrected code (left), multiple tokens (in this case, a pair of indent and dedent) need to be inserted or dropped accordingly, which random perturbations would generate with extremely low probability.
This distribution mismatch between synthetic data and real data results in low performance (\S \ref{sec:experiments}).

To address this challenge, we propose \textit{Break-It-Fix-It} (\textit{BIFI}), an approach that adapts the fixer automatically towards real distributions of bad examples.
Concretely, we first start from the synthetic paired data $\langle$synthetic bad, good$\rangle$ and train an initial fixer as in prior works (see Figure \ref{fig:approach} top left; initialization).
We then perform the following cycle (see Figure \ref{fig:approach} bottom right): (1) we
apply the initial fixer to the real bad examples and use the critic to assess if the fixer's output is good---if good, we keep the pair; (2) we train a \textit{breaker} on the resulting paired data---as this data consists of real code errors, intuitively, the breaker learns to generate realistic code errors; (3) we apply the breaker to the good examples; (4) we finally train the fixer on the newly-generated paired data in (1) and (3).
We iterate this cycle to improve the fixer and the breaker simultaneously in the process.
The intuition is that a better fixer and breaker will be able to generate more realistic paired data, which in turn helps to train a better fixer and breaker.

Below, we describe the initialization step in \S \ref{sec:approach-init}, our main algorithm BIFI in \S \ref{sec:approach-BIFI}, and discuss two baselines of BIFI: backtranslation (\S \ref{sec:approach-BT}) and \textit{FixerOnly} (version of BIFI without the breaker; \S \ref{sec:approach-FixerOnly}).

\subsection{Initialization}
\label{sec:approach-init}
Given unlabeled data $\gD = (\gD_\text{bad}, \gD_\text{good})$, we first prepare a synthetic paired data $\gP_\text{synthetic}$ by perturbing good examples:
\begin{align}
\gP_\text{synthetic} &= \{(b_\text{synthetic}(y),~ y) \mid y \in \gD_\text{good} \},\label{eq:P_syn}
\end{align}
where $b_\text{synthetic}$ denotes a pre-defined procedure that corrupts code.
For instance, we will experiment with two choices of $b_\text{synthetic}$: (i) \textbf{random noising}, which randomly drops \!/\! inserts \!/\! replaces tokens in code, and (ii) \textbf{heuristic noising} designed in \citet{yasunaga2020repair}, which aims to generate common programming errors such as typo, punctuation and type errors. More details are described in \S \ref{sec:exp-setup}.

We then train the initial fixer $f_0$ and breaker $b_0$ on the synthetic paired data:
\begin{align}
b_0 &= \traingb ( \gP_\text{synthetic} )\\
f_0 &= \trainbg ( \gP_\text{synthetic})
\end{align}
where $\traingb(\gP)$ denotes training an encoder-decoder model that maps good-side examples to bad-side examples in a paired data $\gP$, and $\trainbg(\gP)$ does the reverse.
Note that $f_0$ here corresponds to the fixer learned in prior works (\eg, \citet{gupta2017deepfix,yasunaga2020repair}).
We call this initialization step our \textit{round} 0.

\subsection{{Break-It-Fix-It} ({BIFI})}
\label{sec:approach-BIFI}
BIFI aims to improve the fixer and breaker simultaneously through rounds of data generation and training: (1) use a fixer to create data for a breaker, (2) train a breaker, (3) use a breaker to create data for a fixer, and (4) train a fixer.
Concretely, after the initialization step, BIFI performs the following in each round $k$ $=1,2,\dots, K$:
\begin{align}
\gP_k^{(f)} &= \{ (x,~ f_{k\!-\!1}(x)) \mid x \in \gD_\text{bad},~ \textcolor{darkred}{c(f_{k\!-\!1}(x)) = 1} \}\label{eq:BIFI-Pf}\\
b_k &= \traingb ( \gP_k^{(f)} )\label{eq:BIFI-b}\\
\gP_k^{(b)} &= \{ (b_{k}(y),~ y)  \mid y \in \gD_\text{good},~ \textcolor{darkred}{c(b_{k}(y)) = 0} \}\label{eq:BIFI-Pb}\\
f_k &= \trainbg ( \textcolor{darkred}{\gP_k^{(f)}~\cup}~ \gP_k^{(b)} ).\label{eq:BIFI-f}
\end{align}
For convenience, we call the original examples in $\gD_\text{bad}$ \textit{real} bad examples.
Here Eq \ref{eq:BIFI-Pf} applies the current fixer $f_{k-1}$ to the real bad examples in $\gD_\text{bad}$, and keeps outputs that are actually fixed (verified by the critic $c$; \textcolor{darkred}{\textbf{red part}}). This way, we can obtain new paired data $\gP_k^{(f)}$ that is based on \textit{real} bad examples.
Eq \ref{eq:BIFI-b} then trains the breaker $b_k$ (fine-tunes from the previous breaker $b_{k-1}$) on this new paired data $\gP_k^{(f)}$ so that intuitively it can learn to generate realistic bad examples.
Next, Eq \ref{eq:BIFI-Pb} applies the breaker $b_k$ to the good examples in $\gD_\text{good}$, and keeps outputs that are actually broken (verified by the critic $c$; \textcolor{darkred}{\textbf{red part}}). 
This provides an extra paired data $\gP_k^{(b)}$ that is based on bad examples generated by the learned breaker.
Finally, Eq \ref{eq:BIFI-f} trains the fixer $f_k$ (fine-tunes from the previous fixer $f_{k-1}$) on both $\gP_k^{(f)}$ and $\gP_k^{(b)}$, so that the fixer sees real \emph{and} breaker-generated bad examples.
Over time, this cycle adapts the fixer and breaker towards the distribution of real examples.
Figure \ref{fig:approach} (bottom right) provides an illustration of BIFI.

\subsection{Comparison with Backtranslation}
\label{sec:approach-BT}

The BIFI algorithm is related to backtranslation \cite{lample2018phrase} in unsupervised machine translation. 
One may view the bad-side and good-side in our setup as two source \slash target languages in machine translation.
Backtranslation uses a target-to-source model to generate noisy sources and trains a source-to-target model to reconstruct the targets. Specifically, in each round $k$, backtranslation performs the following:
\begin{align}
\gP_k^{(f)} &= \{ (x,~ f_{k\!-\!1}(x)) \mid x \in \gD_\text{bad} \}\label{eq:BT-Pf}\\
b_k &= \traingb ( \gP_k^{(f)} )\\
\gP_k^{(b)} &= \{ (b_{k}(y),~ y)  \mid y \in \gD_\text{good} \}\label{eq:BT-Pb}\\
f_k &= \trainbg (\gP_k^{(b)} )\label{eq:BT-f}.
\end{align}
BIFI differs in two aspects. 
First, as our task has a critic, BIFI uses the critic to verify the outputs of the fixer and breaker, and only keep examples whose outputs are actually fixed (Eq \ref{eq:BIFI-Pf} \textcolor{darkred}{\textbf{red part}}) and whose outputs are broken (Eq \ref{eq:BIFI-Pb} \textcolor{darkred}{\textbf{red part}}) to generate paired data.
In contrast, backtranslation does not verify the generated paired data in Eq \ref{eq:BT-Pf}, \ref{eq:BT-Pb}. This may blindly include non-fixed code as good-side examples (consequently, the breaker might learn to output erroneous code) and non-broken code as bad-side examples, leading to noisy training data (\eg, Figure \ref{fig:approach} bottom left).
Secondly, while backtranslation trains the fixer only on examples {predicted} by the breaker (Eq \ref{eq:BT-f}), BIFI trains the fixer on the \textit{real} bad examples (\textcolor{darkred}{{$\gP_k^{(f)}$}} in Eq \ref{eq:BIFI-f}) in addition to examples generated by the breaker, which improves the correctness and distributional match of training data. 
We will show in our ablation study (\S \ref{sec:analysis-BT}) that both of these two components improve the learning of a fixer and breaker.
In essence, BIFI is an augmentation of backtranslation with a critic.

\subsection{Version of BIFI without breaker: \textit{FixerOnly}}
\label{sec:approach-FixerOnly}
The benefit of BIFI is to enable training the fixer on real bad examples \textit{and} bad examples generated by a learned breaker.
We consider a version (\textit{FixerOnly}) that trains the fixer simply on the real bad examples (prepared in Eq \ref{eq:BIFI-Pf}) but not the bad examples generated by the breaker (Eq \ref{eq:BIFI-Pb}).
Specifically, {FixerOnly} does the following in each round $k$:
\begin{align}
\gP_k^{(f)} &= \{ (x,~ f_{k\!-\!1}(x)) \mid x \in \gD_\text{bad},~ c(f_{k\!-\!1}(x)) = 1 \}\\
f_k &= \trainbg ( \gP_k^{(f)} ).
\end{align}
FixerOnly can also be viewed as self-training \cite{lee2013pseudo} with the difference that we only add fixer outputs verified by the critic to the training data.
We will show in \S \ref{sec:analysis-realbad} that FixerOnly is especially useful when the amount of available bad examples $|\gD_\text{bad}|$ is big, but the gain is smaller compared to BIFI when $|\gD_\text{bad}|$ is small, because BIFI can use the breaker to generate additional paired data for training the fixer (Eq \ref{eq:BIFI-Pb}).

\section{Experiments}
\label{sec:experiments}

We evaluate our approach on two code repair datasets: a common benchmark \textbf{DeepFix}\footnote{{\url{https://bitbucket.org/iiscseal/deepfix}}} \citep{gupta2017deepfix}, and \textbf{GitHub-Python}, a bigger dataset we collect in this paper.

\subsection{Dataset and setup}
\label{sec:exp-setup}

We first describe the detail and experimental setup for GitHub-Python and DeepFix.

\subsubsection{Github-Python}
\heading{Dataset.}
To obtain an unlabeled dataset of code, we collected Python3 files from GitHub public repositories.\footnote{\url{https:github.com}\vspace{0mm}}
We then tokenize each code file using the builtin Python tokenizer, and keep code snippets of length 10--128 tokens, resulting in 3M code snippets.
As the critic $c$, we use the Python AST parser,\footnote{\url{https://docs.python.org/3/library/ast.html}} 
which catches unbalanced parentheses, indentation errors, and other syntax errors. Concretely, we define $c(x) = 1$ (good) if the AST parser returns no errors for input code $x$, and $c(x) = 0$ (bad) otherwise.
Using this critic, we obtain 38K snippets of bad code and 3M snippets of good code.
From the 38K bad examples, we holdout 15K as the final test set, and make the remaining 23K bad examples available for BIFI.
Our goal is to learn a fixer that repairs AST parse errors. We define that the fixer's repair is successful if the output code has no AST parse errors and has Levenshtein edit-distance \cite{levenshtein1966binary} less than 5 tokens from the input code.
The evaluation metric is the fixer's repair accuracy on the test set, \ie, the heldout 15K examples of real bad code.

\begin{table}[!t]
\newcolumntype{G}{>{\columncolor{LightGray}}c}
\hspace{-1mm}
\scalebox{0.85}{
\begin{tabular}{llGccc}
    \toprule  
    \multirow{2}{*}{\textbf{Method}\vspace{-5mm}} & 
    &\multicolumn{4}{c}{\textbf{Test accuracy}}\\
    \cmidrule(lr){3-6}
      & &  
    \!\!\textbf{Total}\!\! &  \!\!\begin{tabular}{@{}c@{}}\scalebox{0.9}{Unbalanced}\\[-1mm]\scalebox{0.9}{Parentheses}\end{tabular}\!\!\! &  \begin{tabular}{@{}c@{}}\!\scalebox{0.9}{Indentation}\!\!\!\\[-1mm]\scalebox{0.9}{Error}\end{tabular} &  \begin{tabular}{@{}c@{}}\scalebox{0.9}{Invalid}\\[-1mm]\scalebox{0.9}{Syntax}\end{tabular}\\
    \midrule  
    {Initial} & 
    Round-0~~~ & \!62.0\%\! & 87.7\% & 39.4\% & 70.5\% \\
    \hdashline[2pt/1.5pt]&\\[-3.7mm]
    \multirow{2}{*}{\begin{tabular}{@{}l@{}}\textbf{FixerOnly}\end{tabular}}\!\! 
    & Round-1~~ & \!86.8\%\! & 93.3\% & 79.5\% & 90.9\% \\
    & Round-2~~~ & \!88.6\%\! & 92.4\% & 83.7\% & 92.0\% \\
    \hdashline[2pt/1.5pt]&\\[-3.7mm]
    \multirow{2}{*}{\begin{tabular}{@{}l@{}}\textbf{BIFI}\end{tabular}}
    & Round-1~~~ & \!88.0\%\! & 94.1\% & 81.3\% & 91.6\%  \\
    & Round-2~~~ & \!\textbf{90.5\%}\! & \textbf{94.2\%} & \textbf{85.9\%} & \textbf{93.5\%}  \\
    \bottomrule 
\end{tabular}
}\vspace{-3mm}
\caption{
\textbf{Repair accuracy on the GitHub-Python test set}. 
The initial fixer is trained on synthetic bad code. 
Our proposed method (BIFI) enables the fixer to be fine-tuned on \textit{real} bad code and bad code generated by the learned breaker. The result shows that BIFI outperforms the initial fixer by a large margin.
}\vspace{-2mm}
\label{tbl:result_python}
\end{table}

\begin{table}[!tb]
\centering
\scalebox{0.85}{
\begin{tabular}{llc}
\toprule
\multicolumn{2}{l}{\textbf{Method}}
& \textbf{Test accuracy}\\
\midrule
\multicolumn{2}{l}{DeepFix \scalebox{0.9}{~\hspace{3.1mm}\citep{gupta2017deepfix}}} & 33.4\% \\
\multicolumn{2}{l}{RLAssist \scalebox{0.9}{~\hspace{2.1mm}\citep{gupta2018deep}}} & 26.6\%\\
\multicolumn{2}{l}{SampleFix \scalebox{0.9}{~\hspace{-0.3mm}\citep{hajipour2019samplefix}}} & 45.3\% \\
\multicolumn{2}{l}{DrRepair \scalebox{0.9}{~\hspace{2.1mm}\citep{yasunaga2020repair}}}~~~ & 66.1\% \\
\midrule
Our Initial & Round-0  ~~(=\,DrRepair)& 66.1\% \\
\hdashline[2pt/1.5pt]&\\[-3.7mm]
\multirow{2}{*}{\begin{tabular}{@{}l@{}}\textbf{Our FixerOnly}\end{tabular}} & Round-1 & 68.6\% \\
& Round-2 & 70.5\% \\
\hdashline[2pt/1.5pt]&\\[-3.7mm]
\multirow{2}{*}{\begin{tabular}{@{}l@{}}\textbf{Our BIFI}\end{tabular}} & Round-1 & 70.8\% \\
& Round-2 & \textbf{71.7\%} \\
\bottomrule
\end{tabular}
}\vspace{-1mm}
\caption{\textbf{Repair accuracy on the DeepFix test set}. 
We define our initial fixer (Round 0) to be the existing best system DrRepair. Note that DrRepair's training procedure coincides with the initialization step of our BIFI algorithm, with heuristic perturbations used in Eq \ref{eq:P_syn}. 
We then apply BIFI on top of it for Round 1, 2. BIFI outperforms DrRepair, achieving a new state-of-the-art.
}\vspace{-2mm}
\label{tbl:result_deepfix}
\end{table}

\heading{BIFI implementation details.}
For the architecture of the fixer and breaker, we use the encoder-decoder Transformer \cite{vaswani2017attention} with 4 layers, 8 attention heads, and hidden states of size 256.
The model parameters are optimized by Adam \cite{kingma2015adam}, with batch size of 27,000 tokens, learning rate 0.001, and gradient clipping 1.0 \cite{pascanu2013difficulty}, on one GPU (GTX Titan X).
For generation, we use beam search with beam size 10,
and keep predictions with Levenshtein edit-distance less than 5 tokens from the input.

To train the initial fixer $f_0$, we use random perturbations for the corruption procedure $b_\text{synthetic}$ (Eq \ref{eq:P_syn}), which drops, inserts, or replaces 1--3 tokens in code with uniform distribution. We apply $b_\text{synthetic}$ 8 times to each of the 2.6M good code snippets to prepare the initial training data $\gP_\text{synthetic}$. We holdout 1\% of $\gP_\text{synthetic}$ as our dev set, which we use to perform early stopping.
We then run the BIFI algorithm for $K = 2$ rounds.

\subsubsection{DeepFix}
\heading{Dataset.}
\label{sec:exp-setup-deepfix}
DeepFix \cite{gupta2017deepfix} contains C code submitted by students in an introductory programming course, of which 37K snippets are good (no compiler error) and 7K are bad (have compiler errors). Each code snippet has 25 lines on average.
Within the 7K bad examples, we take 20\% as a heldout test set. We make the remaining 80\% available for BIFI.
The goal is to learn a fixer that repairs compiler errors. Repair is successful if the output code has no compiler errors. The evaluation metric is the fixer's repair accuracy on the test set.

\heading{BIFI implementation details.}
We define the initial fixer $f_0$ as DrRepair \cite{yasunaga2020repair} (the existing best system on DeepFix), which is an encoder-decoder model trained in a procedure that corresponds exactly to the initialization step of BIFI (\S \ref{sec:approach-init}).
Specifically, to train DrRepair, \citet{yasunaga2020repair} design heuristic perturbations for the corruption procedure $b_\text{synthetic}$, which mimics common code errors beginner and experienced programmers make (\eg, typos, punctuation, keyword and type errors).
We use the same training \!/\! dev data prepared and released by the authors to train the initial fixer. 
We then run the BIFI algorithm for $K=2$ rounds. Our fixer and breaker have the same model architecture as DrRepair. 
At test time, following the original DrRepair, we repeatedly apply the fixer while the code still has errors, up to a maximum of 5 times.

\subsection{Main results}

\begin{table}[!t]
\newcolumntype{G}{>{\columncolor{LightGray}}c}
\centering
\scalebox{0.85}{
\begin{tabular}{llcccc}
    \toprule  
    \multirow{2}{*}{\textbf{Method}\vspace{-5mm}} & 
    &\multicolumn{4}{c}{\textbf{Test accuracy}}\\
    \cmidrule(lr){3-6}
      & &  
    \begin{tabular}{@{}c@{}}\scalebox{0.95}{Bad}\\[-0.7mm]\scalebox{0.9}{100\%}\end{tabular} &  \begin{tabular}{@{}c@{}}\scalebox{0.95}{Bad}\\[-0.7mm]\scalebox{0.9}{50\%}\end{tabular} &  \begin{tabular}{@{}c@{}}\scalebox{0.95}{Bad}\\[-0.7mm]\scalebox{0.9}{10\%}\end{tabular} &  \begin{tabular}{@{}c@{}}\!\!\!\!\scalebox{0.9}{ref. Synthetic}\\[-0.7mm]\scalebox{0.95}{bad only}\end{tabular}\\
    \midrule  
    {Initial} & 
    Round-0~~~ & 62.0\% & 62.0\% & 62.0\% & 62.0\% \\
    \hdashline[2pt/1.5pt]&\\[-3.7mm]
    FixerOnly & Round-2~~~ & 88.6\% & 84.7\% & 78.5\% & 62.7\% \\
    \hdashline[2pt/1.5pt]&\\[-3.7mm]
    BIFI & Round-2~~~ & {90.5\%} & 89.0\% & 86.7\% & 63.3\% \\
    \bottomrule 
\end{tabular}
}
\vspace{-3mm}
\caption{
\textbf{Analysis varying the amount of (real) bad code on GitHub-Python}. ``Bad 100\%'' is our original setting (with 23K real bad examples available for BIFI) and ``Bad 50\%'' means only 50\% of them (11.5K) are made available. ``Synthetic bad only'' means keeping training the fixer on synthetic bad examples. The result shows that (i) even when the amount of real bad examples is small (\eg, ``Bad 10\%''), they are still useful, allowing FixerOnly \slash BIFI to perform better than using synthetic data alone; (ii) when the amount of real bad examples is small, BIFI exceeds FixerOnly by a larger margin, highlighting the usefulness of the bad examples generated by the learned breaker.
}\vspace{-3mm}
\label{tbl:result_analysis_bad_amount}
\end{table}

\begin{table}[!tb]
\centering
\scalebox{0.85}{
\begin{tabular}{llcc}
\toprule
\multicolumn{2}{l}{\textbf{Method}}
& \textbf{Test accuracy}\\
\midrule
Initial 
& Round-0~~\hspace{5mm} & 62.0\% \\
\midrule
BIFI (\textbf{ours}) & Round-2~~ & \textbf{90.5\%} \\[0.3mm]
\hdashline[2pt/1.5pt]&\\[-3.4mm]
~-- real bad & Round-2~~ & {84.6\%} \\[0.3mm]
\hdashline[2pt/1.5pt]&\\[-3.4mm]
~-- critic & Round-2~~ & {84.0\%} \\[0.3mm]
\hdashline[2pt/1.5pt]&\\[-3.4mm]
\begin{tabular}{@{}l@{}}~-- both\\[-0.5mm]
(backtranslation)\end{tabular}\!\! & Round-2~~ & {80.1\%} \\
\bottomrule
\end{tabular}
}\vspace{-2mm}
\caption{
\textbf{Performance comparison with backtranslation} on GitHub-Python. Backtranslation is equivalent to removing two components from BIFI: (i) using the critic to verify fix \slash break attempts (``critic'') and (ii) training the fixer on real bad examples in addition to examples generated by the breaker (``real bad''). The result suggests that both of these components improve the learning of a fixer, and consequently BIFI outperforms backtranslation.
}\vspace{-2mm}
\label{tbl:analysis_backtranslation}
\end{table}

\begin{table*}[!t]
\centering
\scalebox{0.78}{
\small
\begin{tabular}{lrccccc}
    \toprule  
    \multirow{2}{*}{\textbf{Code error category}\vspace{-1mm}} &
    \multirow{2}{*}{\textbf{\#Examples}\vspace{-1mm}} & \multirow{2}{*}{\begin{tabular}{@{}c@{}}\textbf{Initial}\\[-0.mm]\textbf{Round-0 Acc.}\end{tabular}\vspace{-1.5mm}
    } &\multicolumn{2}{c}{ \textbf{FixerOnly}}&\multicolumn{2}{c}{ \textbf{BIFI}}\\
    \cmidrule(lr){4-5} \cmidrule(lr){6-7}
     &  & & \textbf{Round-1 Acc.} & \textbf{Round-2 Acc.} & \textbf{Round-1 Acc.} & \textbf{Round-2 Acc.}\\
    \midrule  
    \rowcolor{Gray}Total  & 15055~~                                & 62.0\% & 86.8\% & 92.4\% & 88.0\% & 90.5\% \\
    \midrule
    \rowcolor{LightGray}Unbalanced Parentheses & 3999~~            & 87.7\% & 93.3\% & 92.4\% & 94.1\% & 94.2\% \\
    ~~~~~Unclosed left parenthesis (not nested)\! & 226~~          & 92.5\% & 94.6\% & 94.6\% & 94.7\% & 94.4\% \\
    ~~~~~Unclosed left parenthesis (nested) & 3014~~               & 85.8\% & 92.8\% & 91.7\% & 93.8\% & 93.8\% \\
    ~~~~~Redundant right parenthesis & 759~~                       & 93.8\% & 95.3\% & 94.7\% & 95.4\% & 95.7\% \\
    \midrule
    \rowcolor{LightGray}Indentation Error & 6307~~                 & 39.4\% & 79.5\% & 83.7\% & 81.3\% & 85.9\% \\
    ~~~~~Expected indent & 4311~~                                  & 46.4\% & 81.2\% & 84.1\% & 82.0\% & 85.3\% \\
    ~~~~~Unexpected indent & 1966~~                                & 24.6\% & 76.8\% & 83.8\% & 80.9\% & 88.4\% \\
    \midrule
    \rowcolor{LightGray}Invalid Syntax  & 4749~~                   & 70.5\% & 90.9\% & 92.0\% & 91.6\% & 93.5\% \\
    ~~~~~Missing colon & 663~~                                     & 98.3\% & 97.3\% & 97.4\% & 98.2\% & 98.0\% \\
    ~~~~~Missing comma (single-line list/tuple/dict)\!\!\! & 694~~ & 95.4\% & 98.1\% & 97.4\% & 98.4\% & 98.3\% \\
    ~~~~~Missing comma (multi-line list/tuple/dict)\!\!    & 451~~ & 88.9\% & 92.5\% & 92.0\% & 94.5\% & 94.9\% \\
    ~~~~~Missing newline & 52~~                                    & 84.6\% & 86.5\% & 88.5\% & 86.5\% & 88.5\% \\
    ~~~~~Missing parenthesis pair & 634~~                          & 82.5\% & 85.0\% & 86.4\% & 87.1\% & 88.3\% \\
    ~~~~~Redundant comma & 152~~                                   & 73.7\% & 84.2\% & 91.4\% & 84.9\% & 92.1\% \\
    ~~~~~Redundant parenthesis pair & 698~~                        & 13.8\% & 80.7\% & 86.1\% & 80.1\% & 89.4\% \\
    ~~~~~\begin{tabular}{@{}l@{}}Invalid use of comma\\[-1.2mm]~~~~~\scalebox{0.7}{(\eg, ``\texttt{raise OSError, \!"msg"}'' $\rightarrow$ ``\texttt{raise OSError("msg")}'')}\end{tabular}\!\!\!\!
    & 1138~~ 
                                                                   & 61.3\% & 98.8\% & 99.1\% & 98.7\% & 99.4\% \\
    ~~~~~Other & 267~~                                             & 60.7\% & 66.3\% & 64.4\% & 67.4\% & 66.7\% \\
    \bottomrule 
\end{tabular}
}
\vspace{-1mm}
\caption{
\textbf{Code error categories in GitHub-Python, and repair accuracy}. 
Due to the mismatch between real errors and synthetic perturbations used for training, the initial fixer has lower accuracy on ``nested'' than ``not nested'' for ``unbalanced parentheses'' errors, but it catches up in BIFI round 1,\,2. Similarly, the initial fixer's repair accuracy is very low for ``redundant parenthesis pair'' and ``indentation error'', but it improves significantly in round 1,\,2.
This result illustrates the effect of BIFI adapting the fixer towards real errors.
See \S \ref{sec:analysis-adapt} for more analysis.
} \vspace{-2mm}
\label{tbl:result_python_detail}
\end{table*}

We study the fixer's repair accuracy on GitHub-Python and DeepFix. Here ``round $k$ accuracy'' means the repair accuracy of the fixer learned in round $k$, \ie, $f_k$.

\heading{GitHub-Python.}
Table \ref{tbl:result_python} shows the test results on GitHub-Python. We show the overall repair accuracy (``Total'') as well as the breakdown over the error categories in the Python AST parser (table right).
The initial fixer $f_0$ (``Initial'') 
is trained on randomly perturbed, synthetic bad code. 
Our proposed method (BIFI) enables the initial fixer to be further trained on real bad code and bad code generated by the learned breaker, which outperforms the initial fixer significantly: +28.5\% in overall repair accuracy, and consistently across all error categories.
This result suggests that even if we start from a very simple initial fixer trained with random perturbations, BIFI can automatically turn it into a usable fixer with high repair accuracy---90.5\% accuracy on real bad code.
We also experimented with continuing training the initial fixer $f_0$ with synthetic perturbations only, for the same rounds of BIFI (hence, controlling the amount of training data seen by the fixer); however, this did not provide an improvement, suggesting that there is a performance ceiling if we only train on synthetic data.

\heading{DeepFix.}
Table \ref{tbl:result_deepfix} shows the test results on DeepFix, along with prior works.
Here we use the existing best system DrRepair as our initial fixer (``Initial''). BIFI outperforms the initial fixer by a substantial margin (+5.6\% absolute over DrRepair), attaining a new state-of-the-art accuracy of 71.7\%.
It is notable that DrRepair was trained with manually-designed heuristic perturbations, where the authors \cite{yasunaga2020repair} mimicked various code errors beginner and experienced programmers make (\eg, typos, punctuation and type errors). Nevertheless, our result suggests that there is still room for improving the adaptation to a more realistic distribution of coding errors, and BIFI boosts repair accuracy without additional manual effort.

\subsection{Analysis}
\label{sec:analysis}

\begin{figure}[!t]
    \hspace{-2mm}
    \includegraphics[width=0.49\textwidth]{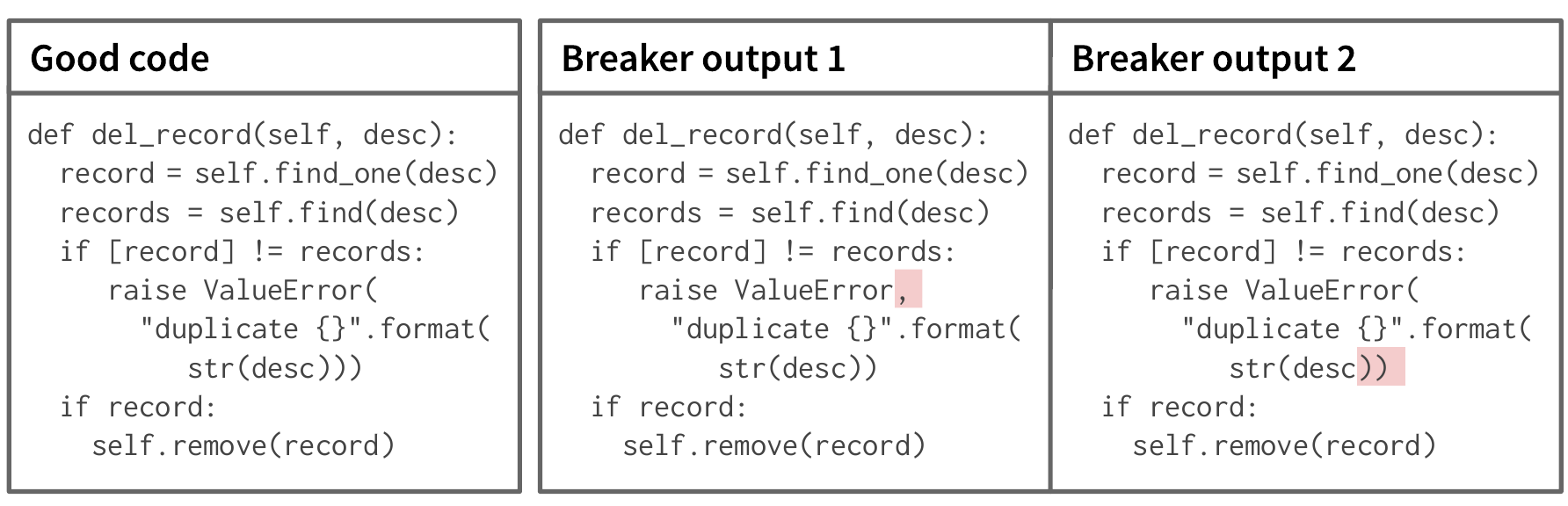}\vspace{-7mm}
    \caption{
    \textbf{Example of breaker outputs}. Given good code on the left, we sampled two outputs made by the breaker learned in BIFI round 1 (right). We observe that the breaker places high probability on errors seen in real bad code (\ie, obsolete usage of \texttt{raise}, unbalanced parentheses in nested context).
    } \vspace{-3mm}
  \label{fig:breaker_pred}
\end{figure}

\begin{figure}[!t]
    \hspace{-2mm}
    \includegraphics[width=0.49\textwidth]{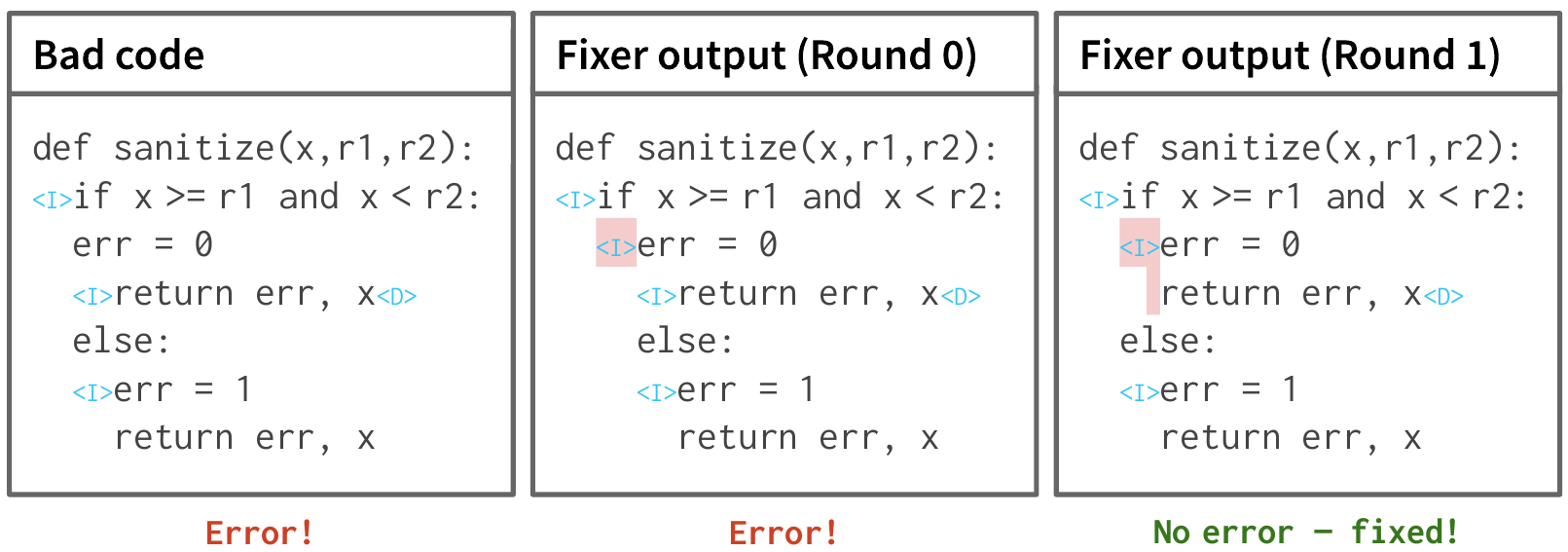}\vspace{-7mm}
    \caption{
    \textbf{Example of fixer outputs}. Given the bad code on the left (with an indentation error), the initial fixer (center) attempts to fix it by inserting an indent token (\texttt{$\langle$I$\rangle$}) to line 3, but fails to adjust (delete) the indent token on the next line. The initial fixer commonly makes this mistake due to the distribution mismatch between real errors and synthetic perturbations on which the initial fixer was trained (see \S \ref{sec:analysis-adapt}). After one round of BIFI, the fixer in round 1 (right) learns to insert and delete the correct pair of indent tokens, fixing the error.
    } \vspace{-2mm}
  \label{fig:fixer_pred}
\end{figure}

We now analyze the key insights of BIFI. As the main properties of BIFI are to (i) add real bad examples to the training data if the critic accepts the fixer's output (the FixerOnly version), and to (ii) train the breaker to generate realistic bad examples (BIFI), we analyze their effects in \S \ref{sec:analysis-realbad} and \S \ref{sec:analysis-breaker}.
We also compare with backtranslation in \S \ref{sec:analysis-BT}.
We then analyze how our fixer adapts towards real code distributions through quantitative and qualitative studies (\S \ref{sec:analysis-adapt}).

\subsubsection{Effect of real bad examples}
\label{sec:analysis-realbad}

FixerOnly enables training the fixer on real bad examples (but does not use the bad examples generated by the breaker). As Table \ref{tbl:result_python} and \ref{tbl:result_deepfix} show, FixerOnly outperforms the initial fixer by a large margin, \eg, +27\% on GitHub-Python.
This result highlights the importance of training on real bad examples.

We further analyze the effect of varying the \textit{amount} of real bad examples, shown in Table \ref{tbl:result_analysis_bad_amount}. Here, ``Bad 100\%'' is our original setting (with 23K real bad examples available for BIFI and FixerOnly) and ``Bad 50\%'' means only 50\% of them (11.5K) are made available. ``Synthetic bad only'' means keeping training the fixer on synthetic bad examples only. We find that while the repair accuracy drops as we decrease the amount of available real bad examples, a small amount of real bad examples (\eg, ``Bad 10\%'') is still useful, making FixerOnly perform better than using synthetic data alone (\ie, 78.5\% vs 62.7\%).

\subsubsection{Effect of bad examples generated by breaker}
\label{sec:analysis-breaker}

Recall that BIFI trains the fixer on both real bad examples and bad examples generated by the learned breaker. As Table \ref{tbl:result_python} and \ref{tbl:result_deepfix} show, BIFI consistently provides an extra boost over FixerOnly, suggesting that the use of breaker outputs improves the fixer. Moreover, another benefit of the breaker is that one can sample many bad examples from the breaker to augment real bad examples, if their amount is limited.
In Table \ref{tbl:result_analysis_bad_amount} we find that BIFI is especially stronger than FixerOnly when the amount of available real bad examples is small (\eg, ``Bad 10\%'').

Figure \ref{fig:breaker_pred} shows sample outputs made by the learned breaker $b_1$ given the good code on the left. We observe that the breaker places high probability on errors seen in real bad code, \ie, obsolete usage of \texttt{raise} in Python3 (center) and unbalanced parentheses in nested context (right). Compared to random perturbations that arbitrarily drop \slash insert tokens, the learned breaker improves the coverage and efficiency of the training data for the fixer.

\subsubsection{Comparison with backtranslation} 
\label{sec:analysis-BT}

Table \ref{tbl:analysis_backtranslation} compares our method (BIFI) with backtranslation.
As discussed in \S \ref{sec:approach-BT}, backtranslation is equivalent to removing two components from BIFI: (i) using the critic to verify fix \slash break attempts in data generation (``critic'' in Table) and (ii) training the fixer on real bad examples besides examples generated by the breaker (``real bad''). We find that removing each component from BIFI hurts the performance (\eg, 90\%$\rightarrow$84\%), which suggests that both components are important to improve the quality of training data. With these two innovations, BIFI outperforms backtranslation by a large margin (+10\% absolute on GitHub-Python).

\subsubsection{How does the fixer adapt?}
\label{sec:analysis-adapt}

We take a closer look at how our fixer performs and adapts towards the real distribution of bad code. Table \ref{tbl:result_python_detail} shows fine-grained categories of code errors seen in GitHub-Python, and their repair accuracy.

In this categorization, we observe two examples of distribution mismatch between the real errors and synthetic perturbations: (i) random perturbations can generate this category of errors with high probability but with a wrong ``sub-distribution'' within it (\eg, can generate ``unbalanced parentheses'' or ``missing comma'' errors but do not match the low-level distribution of real bad code, such as errors occurring more often in nested parentheses or in a multi-line list/tuple/dict; recall the Figure \ref{fig:challenge} top example); 
(ii) random perturbations can only generate this category of errors with very small probability (\eg, ``redundant parenthesis pair'' and ``indentation error'', for which an \textit{exact pair} of parentheses or indents/dedents need to be dropped or inserted; recall the Figure \ref{fig:challenge} bottom example).
For (i), the result shows that the initial fixer trained with random perturbations has lower accuracy on ``nested'' than ``not nested'' for ``unbalanced parentheses'' errors, and on ``multi-line'' than ``single-line'' for ``missing comma'' errors; but the performance catches up in round 1,\,2, suggesting the effect of BIFI for addressing the low-level distribution mismatch.
For (ii), the result shows that the initial fixer's repair accuracy is very low for ``redundant parenthesis pair'' and ``indentation error'', but it achieves significantly higher performance in round 1,\,2 (\eg, 39\%$\rightarrow$85\% for indentation errors). A possible explanation is that as BIFI iteratively adds the successfully repaired cases to training, the fixer adapts to up-weight this category of error fixing, leading to improved accuracy.

Figure \ref{fig:fixer_pred} provides examples of fixer outputs. Given the bad code on the left (with an indentation error), the initial fixer (center) attempts to fix it by inserting an indent token (\texttt{$\langle$I$\rangle$}) to line 3 but fails to adjust (delete) the indent token on the following line. The initial fixer commonly makes this mistake for indentation errors, due to the mismatch between real errors and synthetic perturbations discussed above. After one round of BIFI, the fixer (Figure \ref{fig:fixer_pred} right) learns to insert and delete the correct pair of indents, fixing the error.

\section{Related work and discussion}

\heading{Learning to repair code.}
Several works learn to repair code from labeled datasets of source code edits made by programmers, \eg, error resolution records  \citep{just2014defects4j,chen2019sequencer,mesbah2019deepdelta,bader2019getafix,tarlow2020learning,ding2020patching}.
As labeled data is costly to prepare, various works study learning code fixers from unlabeled or synthetic data \cite{pu2016sk_p,parihar2017automatic,ahmed2018compilation,pradel2018deepbugs,wang2017dynamic,vasic2019neural,gupta2018deep,hajipour2019samplefix,hellendoorn2020global}. In particular, \citet{gupta2017deepfix} is an early work that randomly perturbs good code to prepare a synthetic paired data and trains a seq2seq neural network model as a fixer. 
\citet{yasunaga2020repair} improve on it by designing heuristic perturbations that mimic common errors made by programmers.
Different from the above work, our method (BIFI) adapts a naive fixer trained with synthetic errors towards real errors without manual, domain-specific effort.
For a more comprehensive review of automatic code repair, we refer readers to \citet{repair_review}.

\heading{Denoising autoencoding.}
Denoising autoencoding \cite{vincent2008denoise} trains a model that recovers original data from randomly corrupted versions, and is widely used as an effective self-supervised representation learning and pre-training strategy, \eg, in computer vision \cite{erhan2010does} and natural language processing (NLP) (\citet{lewis2019bart}, which randomly drops \slash replaces tokens in a sentence and learns to recover). Our initial fixer trained with random perturbations is a denoising autoencoder. Crucially, instead of using it purely for representation learning, we show that through the BIFI algorithm, one can turn the vanilla denoising autoencoder into a usable fixer that repairs real-world bad examples with high accuracy. On a related note, \citet{lee2019learning} adapt a denoising autoencoder into an autocomplete system.

\heading{Domain adaptation.}
Domain adaptation aims to address the mismatch in data distribution between training and test domains \cite{daume2006domain,quionero2009dataset,kohsagawa2020wilds}. Such domain shifts typically occur across related but different datasets 
\cite{torralba2011unbiased,fang2013unbiased,venkateswara2017deep,yu2018spider,yu2019sparc,peng2019moment,kamath2020selective}, as well as from synthetic data to real data (including sim2real) \cite{wang2015building,ganin2015unsupervised,richter2016playing,peng2018visda,hellendoorn2019code,xu2020autoqa,bellemare2020autonomous}, as synthetic data can be easier to obtain than real data. 
In our repair task, unpaired real data (code snippets on GitHub) is available but \textit{paired} real data is costly to obtain. Hence we considered adaptation from synthetic paired data to real paired data.
Within domain adaptation, our task is also related to the setting where unlabeled data in the test domain is available \cite{sun2016deep,hoffman2018cycada,sun2019test} (in our case, real bad code).
The difference is that as our outputs are structured, we have a critic to check if the fixer's output on the unlabeled data is correct. Leveraging this property, BIFI takes the correct outputs to create training data in the test domain (Eq \ref{eq:BIFI-Pf}); and trains a breaker to generate more data that simulates the test domain (Eq \ref{eq:BIFI-Pb}).

\heading{Data augmentation.}
Data augmentation aims to generate extra training data. 
A common approach is to increase the source side data, for instance by adding modifications or sampling from generative models \cite{hannun2014deep,jia2016data,krizhevsky2017imagenet,antoniou2017data,yasunaga2017robust,yu2018syntax,lee2019learning,berthelot2019mixmatch,xie2019unsupervised}.
Several works also study target side augmentation, which keeps multiple (valid) target predictions made by a model and adds to training. This is commonly used in structured generation problems such as semantic parsing, program synthesis and molecule generation \cite{liang2013learning,berant2013semantic,guu2017language,min2019discrete,zhong2020grounded,yang2020improving}. While our method also augments training data, it differs in two aspects: 1) we use the fixer and breaker to augment both the source and target sides; 2) our goal is not only to increase the amount of training data, but also to adapt to the distribution of interest (real bad examples).

\heading{Self-training.}
Self-training \cite{lee2013pseudo,mcclosky2006effective,kumar2020understanding,xie2020n} applies a trained model to unlabeled data, obtains predicted targets (pseudo-labels), and uses them as extra training examples. Similarly, co-training \cite{blum1998combining} and tri-training \cite{zhou2005tri} train multiple models and add predicted targets on which these models agree. Our method also applies trained models (breaker and fixer) to unlabeled data to generate targets, with a difference that we use the critic to verify the predictions and only keep correct ones.

\heading{Unsupervised machine translation (MT).}
Unsupervised MT learns translators from unpaired corpora \cite{artetxe2018unsupervised,artetxe2017unsupervised,lachaux2020unsupervised}. Backtranslation \cite{sennrich2015improving,lample2017unsupervised,lample2018phrase} is a common approach that uses the target-to-source model to generate noisy sources and then trains the source-to-target model to reconstruct the targets (also related to cycle-consistency \cite{zhu2017unpaired,hoffman2018cycada} and style transfer \cite{shen2017style,yang2018unsupervised,zhang2019style}).
One may view the ``bad-side'' and ``good-side'' in our repair task as two source \slash target languages in MT and apply backtranslation. 
The main difference is that the repair task has a critic, which motivated our BIFI algorithm: BIFI (i) uses the critic to verify if the generated examples are actually fixed or broken (Eq \ref{eq:BIFI-Pf}, \ref{eq:BIFI-Pb}), and (ii) trains the fixer on \textit{real} bad examples besides examples generated by the breaker (Eq \ref{eq:BIFI-f}). We found these techniques improve the correctness of the generated training data (\S \ref{sec:analysis-BT}).
While we focused on code repair in this work, we hope that BIFI can be applied to unsupervised MT and style transfer by introducing a human-based or learned critic.
\section{Conclusion}
We considered the problem of learning a fixer from unpaired data and a critic (code analyzer or compiler), and proposed a new approach, Break-It-Fix-It (BIFI). 
The idea of BIFI is to train a breaker and use the critic to amass more realistic and correct paired data for training the fixer.
Using two code repair datasets (GitHub-Python and DeepFix), we showed how BIFI adapts baseline fixers towards realistic distributions of code errors, achieving improved repair performance.
We note that BIFI is not about simply collecting more training data, but rather turning raw unlabeled data into \textit{usable} paired data with the help of a critic.
This is a potentially powerful and general framework applicable to many areas such as molecular design, text editing, and machine translation.

\section*{Acknowledgments}
We thank Michael Xie, members of the Stanford P-Lambda, SNAP and NLP groups, as well as our anonymous reviewers for valuable feedback.
This work was supported in part by Funai Foundation Fellowship and NSF CAREER Award IIS-1552635.

\section*{Reproducibility}
Code and data are available at \url{https://github.com/michiyasunaga/bifi}.
Experiments are available at \url{https://worksheets.codalab.org/worksheets/0xfddb2ef01a9f4dc0b5d974a5a97174be}.

\bibliography{main}
\bibliographystyle{icml2021}





\end{document}